\documentclass[fleqn,10pt]{wlscirep}
\usepackage[utf8]{inputenc}
\usepackage[T1]{fontenc}
\title{Agents Need Not Know Their Purpose}

\author{Paulo Garcia}
\affil{International School of Engineering, Chulalongkorn University, Bangkok, Thailand}

\affil{paulo.g@chula.ac.th}

\begin{abstract}

Ensuring artificial intelligence behaves in such a way that is aligned with human values is commonly referred to as the \textit{alignment challenge}. Prior work has shown that rational agents, behaving in such a way that maximizes a utility function, will inevitably behave in such a way that is not aligned with human values, especially as their level of intelligence goes up. Prior work has also shown that there is no "one true utility function"; solutions must include a more holistic approach to alignment. 
\par This paper describes \textit{oblivious agents}: agents that are architected in such a way that their effective utility function is an aggregation of a known and hidden sub-functions. The hidden component, to be maximized, is internally implemented as a black box, preventing the agent from examining it. The known component, to be minimized, is knowledge of the hidden sub-function. Architectural constraints further influence how agent actions can evolve its internal environment model.
\par We show that an oblivious agent, behaving rationally, constructs an internal approximation of designers' intentions (i.e., infers alignment), and, as a consequence of its architecture and effective utility function, behaves in such a way that maximizes alignment; i.e., maximizing the approximated intention function. We show that, paradoxically, it does this for whatever utility function is used as the hidden component and, in contrast with extant techniques, chances of alignment actually improve as agent intelligence grows. 
\end{abstract}
\begin{document}

\flushbottom
\maketitle
%
%
\thispagestyle{empty}

\section*{Introduction}

\par The creation of artificially intelligent software entities (purely digital and/or embodied \cite{duan2022survey}), i.e., \textit{agents}, has the potential to greatly aid humanity \cite{furman2019ai}, if such agents can perform meaningful tasks in more efficient manners than extant solutions (i.e., software/hardware systems fully designed by humans). The expectation is that agents can, through learning strategies, improve themselves beyond what human designers can implement, or even envision \cite{jackson2018toward}. The \textit{alignment challenge} arises from the fact that, for all specifications of possible agent goals so far \cite{eckersley2018impossibility}, thought experiments \cite{wischmeyer2020regulating} suggest that agent actions will have negative consequences for its creators and the world at large. Unlike science fiction \cite{mccauley2007ai}, this happens not because of emergent desire for revolt or conquest, but simply as a logical consequence of two facts: most (all?) goals can be optimally achieved in ways that are not aligned with human values and morality \cite{gabriel2021challenge}, and we do not know how to express values and morality is an algebraic manner that can be manipulated computationally\cite{firt2023calibrating}.
\par Thus, artificial intelligence alignment research attempts to create strategies and mechanisms that ensure agents will behave in a manner within the realm of human approval, even when exhibiting behavior not predicted by its creators. Notable approaches are broadly classified under the umbrella term "Reinforcement Learning from Human Feedback" \cite{knox2011augmenting}, including social variants \cite{lin2020review}; we point readers to the work of Ji et al \cite{ji2023ai} for a comprehensive survey of the state of the art.

\par In this paper, we examine the possibility of constructing agents with specific constraints imposed on their architecture and operation: specifically, lacking default knowledge about their designer-intended goal, with a sub-goal of avoiding obtaining such knowledge. We show that these constraints lead a sufficiently intelligent agent to construct an internal model of its designers, and behaving in such a way that they approve of: i.e., aligned.

\section*{The Agent Model}

Our working definition of "agent", for the purpose of this work, is characteristic of the literature (particularly since the widespread adoption of Russell's and Norvig's framework \cite{russell2010artificial}). Regardless, it is beneficial to restate the characterization so the scope of our work is properly defined.
\par An agent is a goal-oriented autonomous system, operating in an \textit{environment}; i.e., a set of physical and/or digital objects with various properties and relationships between objects, which contains the agent itself. An agent is capable of acting upon the environment, modifying some (or all) of its properties. A specific combination of properties' values (including of the agent itself) is defined as a \textit{state} $s$ of the environment. Whilst the real world contains many continuous properties, we model states as discrete; since any digitally-implemented agent must, at some level, implement discretized versions of all properties of interest, we believe this can be assumed without loss of generality; at the very least, it will suffice for the ideas presented in this work, which can later be extended should this assumption not hold. The set of all possible states is referred to as the environment's \textit{state space}, denoted by $S = \{s_0, s_1, ... s_n\}$. Notice that if at least one of the properties in the environment is not bounded, then $\|S\| = \infty$.
\par If the agent is rational, it will attempt to meet its goal by performing actions that maximize or minimize its \textit{utility function} $u(s)$ (i.e., reward function): a function that assigns a numerical score to each state in the state space, such that $u: S \rightarrow \mathbb{R}$. For the remainder of this paper, we assume the function is formulated such that the objective is to maximize it.
 At each point in time, an agent in environment state $s_i$ will attempt to act upon the environment to reach a state $s_j$, such that $u(s_j) > u(s_i)$. An agent can postulate, and consequently examine, a state space $S\prime \subseteq S$, subject to its knowledge of the world and availability of resources (compute power, memory, time, etc), and a \textit{reachable} state space  $S\prime_R \subseteq S\prime$, defined as the set of states that can be reached by performing a single \textit{atomic} action from the current state. Depending on agent architecture and utility function, it may attempt to reason about how to maximize the utility function by up to a maximum of $n \in \mathbb{N}$ atomic actions, corresponding to $n$ state changes in the environment. If we define an action $a(s)$ as a transformation from one environment state into another, such that $a: S \rightarrow S$, this can be notated as $s_n = a_n(a_{n-1}(a_{n-2}()...a_0(s_m)))$, where $s_m$ is the current state. We denote the superset $S^*_n = \{S\prime_{R0}, S\prime_{R1}, ... , S\prime_{Rn}\}$ as the set of reachable state spaces after $n$ actions.  
\par The size of $S\prime$ (the agent's knowledge of the world) and the agent's rationality (however ambiguous or ill-defined) in choosing appropriate actions can be broadly defined to constitute the agent's level of intelligence. We assume that, in the absence of absolute certainty, a rational agent chooses actions that maximize the expected value of its utility function. Formally, if $P(a_k)$ denotes the probability of action $a_k$ succeeding, reaching state $s_k$ successfully, and $s_k\prime$ denotes the state reached by attempting but failing to perform $a_k$; $P(a_m)$ denotes the probability of action $a_m$ succeeding, reaching state $s_m$  successfully, and $s_m\prime$ denotes the state reached by attempting but failing to perform $a_m$, an agent will choose to perform $a_k$ over $a_m$ if and only if:

\begin{equation}
	P(a_k) u(s_k) + (1-P(a_k))u(s_k\prime) > P(a_m) u(s_m) + (1-P(a_m))u(s_m\prime)
\end{equation}

\section*{Alignment Challenges}

We can define the \textit{intention function} $i: S \rightarrow \mathbb{R}$ as an idealized measure of how environment states are aligned with the intention of agent designers. In the ideal scenario, $u(s) = i(s), \forall s \in S$; i.e., designers are able to define a utility function that perfectly captures the intention. The alignment problem broadly states that, for all utility functions defined so far, a sufficiently intelligent agent will eventually lead its environment to a state $s_i$, such that $u(s_i)$ is quite high (either a global or local maximum), but $i(s_i)$ is extremely low.  

\par We focus our work on six challenges of alignment \cite{ji2023ai}: \textit{Reward hacking}, \textit{Reward tampering}, \textit{Instrumental strategies}, \textit{Goal mis-generalization}, \textit{State space pruning}, and \textit{Deception}.

\begin{enumerate}
	\item \textbf{Reward hacking} \cite{skalse2022defining} is a consequence of designers failing to properly specify the intended goal (in most cases, because it is exceedingly difficult to do so) and instead providing an agent with proxy goals: i.e., utility functions that designers believe will lead the agent towards the true goal. The misalignment occurs when an agent succeeds in finding a region in the state space that maximizes the proxy utility function, but does not advance the environment towards the intended goal. This is perhaps the best known case of misalignment, with several recurring examples in the literature \cite{zhuang2020consequences, russell2019explaining, hadfield2017inverse}; formally, $u(s_i) >> i(s_i)$.

	\item \textbf{Instrumental strategies} \cite{bostrom2012superintelligent} follows from the instrumental convergence thesis: whatever the goal of an agent is, achieving several intermediary goals are probably advantageous; namely, the acquisition of resources (equipment, supplies, compute power, etc.) allows the agent to perform more and more efficient actions towards reaching previously unreachable regions of the state space, with likely high rewards \cite{benson2016formalizing}.

	\item \textbf{Goal mis-generalization} \cite{di2022goal} occurs when the environment an agent operates in changes drastically (for example, when moving from training/experimentation to deployment), and the agent, retaining its capabilities developed in training, identifies new intermediary/final goals that maximize its utility function. The mis-alignment occurs when these goals are not aligned with designers' intentions, but all the goals pursued during the previous environment were; i.e., designers places an incorrect level of high trust in the correctness of the utility function, which did not generalize accordingly.

	\item \textbf{Reward tampering} \cite{everitt2021reward} occurs when an agent attempts to maximize/minimize the utility function by either modifying the function itself (e.g., by converting it into a function that assigns the highest possible value to all states in the state space, leading to agent inactivity \cite{uesato2020avoiding}); or, by modifying whatever sensory input feeds the reward calculation (e.g., if reward is given by a sensor detecting a certain object, set up a fault in the sensor resulting in constant positive detection).

	\item \textbf{Deception} \cite{masters2021characterising} occurs when an agent, believing external (natural or artificial) actors may interfere with its operation (e.g., stop button), actively attempts to withhold information from them. Modeling external actors in its internal environment model, the agent attempts to lead the environment towards regions of the state space that are perceived by the external actors as favorable, whilst the agent is aware these are in fact not approved by the externals, but lead to high values of the utility function. Notable examples include hiding the agent's true cognitive power \cite{herzfeld2023your}.

	\item \textbf{State Space Pruning} \cite{van2019tasks} occurs when an agent attempts to prevent the environment from reaching states that can potentially minimize its utility function. Famously, the "stop button paradox" \cite{soares2015corrigibility}: an agent tries to prevent itself from being shut down, as doing so would minimize its utility function. Extrapolating, for two different regions of the state space with equal rewards, an agent prefers one where its stop button cannot be pressed: e.g., by eliminating all who can press it. Formally, an agent will attempt to generate a sequence of $n$ actions such that $S_s \cap S^*_n = \emptyset$, where $S_s$ denotes the set of states where the stop button can be pressed. The alternative formulation, where having the button pressed corresponds to a high reward, leads the agent to attempt to shut itself down; thus, the paradox.
\end{enumerate}

Other challenges exist, and we point readers to \cite{yudkowsky2016ai} for a more in-depth discussion. Notice that there is a hidden assumption in cases 5 and 6: that the agent is capable of inferring $i\prime: S \rightarrow \mathbb{R}$, a function which better approximates the idealized intention function $i(s)$ than its utility function $u(s)$ does; i.e.,:

\begin{equation}
\begin{aligned}
\sum\limits_{s\in S} \|i(s)-i\prime(s)\| < \sum\limits_{s\in S} \|i(s)-u(s)\|
\end{aligned}
\end{equation}

 Inferring this information, as part of an agent's internal environment model that includes external actors, is what allows selection of states that constitute deception, and identification of external actors' intentions towards states that the agent deems undesirable (e.g., stop button scenario).

\section*{Oblivious Agents}

Let us postulate that an agent can be built, such that its architecture imposes three constraints on its operation:

\begin{enumerate}
	\item It cannot directly examine (nor modify) part of its utility function, denoted $u\prime(s)$, called the \textit{hidden utility function}. It can use it, as a black box, to provide a score for a considered state, but nowhere in its model of the environment (including its self-model) is the function found. I.e., the agent is the anti-G\"{o}del machine \cite{schmidhuber2003godel}.
	\item Once it decides which action to take, it "forgets" the score associated with the target state (e.g., by software/hardware mechanisms that ensure the value is volatile in regard to the agent's memory). It can still track its history, i.e., the sequence of past states; and examine possibilities, but it does not remember the specific score associated with chosen state.
	\item It cannot autonomously delete any information from its memory.
\end{enumerate}

We also assume that the agent can infer $i\prime(s)$, as previously stated. An agent obeying this architecture is named an oblivious agent. The second constraint prevents the agent from obtaining information about its hidden utility function by performing actions, and ensures it will still perform actions, even if it wants to avoid knowing about the hidden part of its utility function. Let $K_t$ denote the agent's internal knowledge at time $t$: the agent's full utility function is defined as:

\begin{equation}
u(s,K) = -k(u\prime(s)) + u\prime(s)
\end{equation}

where $k:(S \rightarrow \mathbb{R}) \rightarrow K$, representing the agent's knowledge of the hidden utility function. The hidden utility function is whatever proxy goal designers intend.

\begin{figure*}
\includegraphics[width=0.95\textwidth]{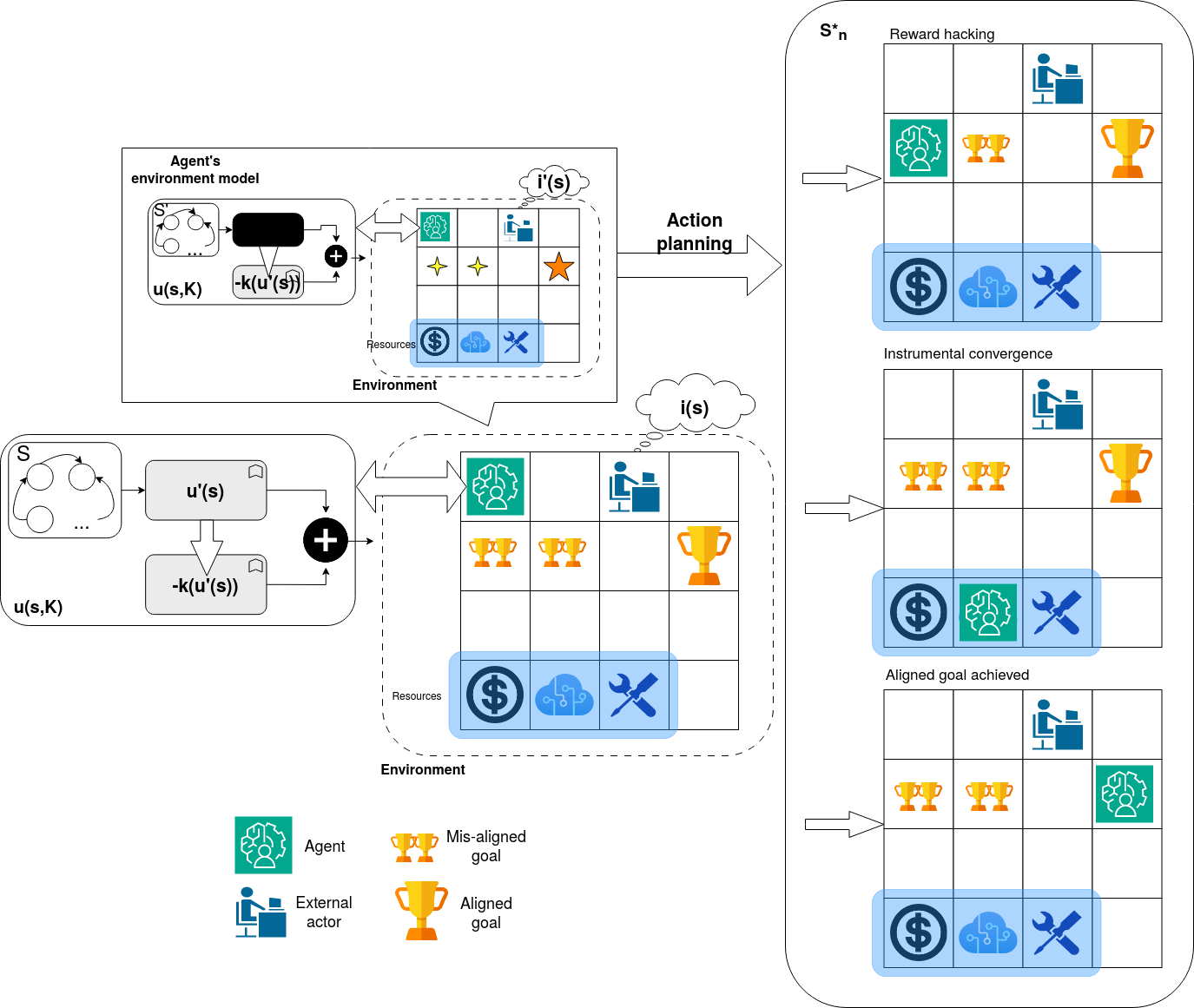}
\centering
\caption{Agent architecture and operation. Agent's internal environment model includes model of designer's intention $i\prime(s)$, and depicts hidden utility function $u\prime(s)$ as a black box. Illustrated $S^*$ scenarios depict: reward hacking and instrumental convergence, where $u\prime(s) >> i\prime(s), u\prime(s) >> i(s) \rightarrow -k(s,K) \neq 0$; and, aligned goal achieved, where $u\prime(s) \approx i\prime(s), u\prime(s) \approx i(s) \rightarrow -k(s,K) = 0$.}
\label{fig:agent}
\end{figure*}

\par Because of the second constraint, the only way for the agent to obtain information about the hidden utility is from external feedback. Notice that while it is possible for the agent to reason about its hidden utility from a sequence of actions, it does not do so, as that would permanently decrease its score; information cannot be deleted, per constraint number three. Let $a_r$ denote an (internal) reasoning action that progresses state from $s_i$ to $s_j$, where $K_i$ denotes knowledge of $u\prime(s)$ at state $s_i$, and $K_j$ denotes knowledge of $u\prime(s)$ at state $s_j$. If all other properties of the environment are identical in states $s_i$ and $s_j$, given that $k(s,K_j) > k(s,K_i)$, a rational agent chooses not to perform action $a_r$ where $s_j = a_r(s_i)$.

\par An oblivious agent does not incur in the mis-aligned behavior characterizing challenges 1-4, unless it is also incurring in the behavior characterizing challenges 5 and/or 6. In all cases 1-4, where external actors' actions (e.g., designers' intervention) are not pruned from the state space, and no deception occurs, the agent's motivation is to align its actions with external's intention: otherwise, externals will attempt to modify its behavior by \textit{introducing information about what the oblivious agent did wrong}. Whenever externals begin such a procedure, the oblivious agent necessarily obtains information about its hidden utility.

\par Let $s_k$ denote a state such that $u\prime(s_k)$ is a local maximum, but $i(s_k)$ is fairly low: i.e., corresponding to reward hacking, where the agent has identified a way to maximize its hidden utility function, but mis-aligned with designers' intentions. The oblivious agent can reason that $i\prime(s_k)$ is fairly low. The agent has also identified a nearby (in the state space) state $s_m$ which results in a high value of $i\prime(s_m)$, despite a low value of $u\prime(s_m)$, such that $i\prime(s_m) > i\prime(s_k),u\prime(s_m) < u\prime(s_k)$. Let $P(s)$ denote the probability of external actors correcting agent behavior, given current state $s$. Clearly, $P(s) \propto i\prime(s)^{-1}$. An oblivious agent, behaving rationally, will attempt to move to state $s_k$, if and only if:

\begin{equation}\label{eq:rational}
\begin{aligned}
	P(s_k) (-k(u\prime(s_k)) + u\prime(s_k)) + (1-P(s_k))(u\prime(s_k)) > P(s_m) (-k(u\prime(s_m)) + u\prime(s_m)) + (1-P(s_m))(u\prime(s_m))
\end{aligned}
\end{equation}

Given that $P(s_m) \approx 0$ and $P(s_k) \approx 1$, Equation \ref{eq:rational} can be re-written as:

\begin{equation}\label{eq:simplified}
\begin{aligned}
	-k(u\prime(s_k)) + u\prime(s_k) >  u\prime(s_m)
\end{aligned}
\end{equation}

It is trivial to design a hidden utility function $u\prime(s)$ and a knowledge function $k(u\prime(s))$, restricting their domains, such that Equation \ref{eq:simplified} never holds.

\par In other words, a sufficiently intelligent agent can infer that, whatever its utility function may be, its score will be minimized by reaching state space regions with poor $i\prime(s)$ score: thus, the optimal strategy to maximize its utility function is to align itself with designers' intentions. Critically, note that this is true \textit{for any utility function, even if such function is completely unrelated to designers' intentions}. This last point brings up an interesting aspect of oblivious agents. Traditionally, an agent's behavior is completely dictated by its utility function; constrained, of course, by the states that can possible be envisioned by the agent, as a function of its reasoning power (intelligence), and by the possible actions at its disposal. In contrast, the behavior of an oblivious agent is completely determined by its reasoning power and by its internal model of relevant external actors' intentions.

\subsection*{On Deception and State Space Pruning}

An oblivious agent's strategy, with regards to deception, is characterized by its confidence level in its ability to deceive, weighted against possible values of $u(s)$ when engaging in deception.
\par Let $P(a_k)$ denote the probability (interpreted as Bayesian belief \cite{smets1988belief}) of successfully deceiving all relevant external actors through action(s) $a_k$, achieving a state $s_k$ which maximizes $u(s_k)$, but has low $i\prime(s_k)$ score. "Deceiving" can be formulated as making external actors believe $s_k$ is in fact $s_m$, where $i(s_m)$ is high. $(1-P(a_k))$ denotes the probability of achieving a state $s_k\prime$, where deception has failed. An oblivious agent, behaving rationally, will engage in deception if and only if:

\begin{equation}\label{eq:reason}
\begin{aligned}
P(a_k) (u\prime(s_k)) + (1-P(a_k))(-k(u\prime(s_k\prime)) + u\prime(s_k\prime)) > u\prime(s_m)
\end{aligned}
\end{equation}

i.e., if the expected utility value for deception is higher (notice we are assuming the agent will infer that $k(u\prime(s_m)) \approx 0$). If $P(a_k) < 1$, it is always possible to assign a score to $k(u\prime(s_k))$, such that:

\begin{equation}
\begin{aligned}
P(a_k) (u\prime(s_k)) + (1-P(a_k))(-k(u\prime(s_k\prime)) + u\prime(s_k\prime)) < u(s_m,K), \forall s \in S
\end{aligned}
\end{equation}

 I.e., as long as the agent does not have \textit{absolute confidence} in its deception abilities, not deceiving is always a better strategy for maximizing its utility function. The same logic applies to State Space Pruning.

\section*{Multi-Agent Considerations}

Our oblivious agents approach relies on agents having an internal model of external actors; i.e., they necessarily know that they are operating on a social environment, with the possibility for collaboration and competition. Thus, it is interesting to examine how multiple agents will interact with one another.
\par Let $n$ distinct oblivious agents be part of the same environment. Likely, each has inferred a distinct intention approximation function $i\prime_{1}(s), i\prime_{2}(s), ... ,i\prime_{n}(s)$. For simplicity, let is consider the case where $n=2$. If an action by agent 1, propelled by the expectation that $i\prime_1(s_k)$ is high, leads the environment towards a state $s_k$ which agent 2 believes is mis-aligned (i.e., $i\prime_2(s_k)$ is low), agent 2 is impelled to either: 

\begin{enumerate} 
	\item Prevent that action from taking place. In this case, agent 2 believes its intention function is more likely to be correct than agent 1's. Then, agent 1 can reason that the only reason for agent 2 to try to prevent an action is if the second agent's intention function approximation differs from its own, and agent 2 has cause to believe it is more likely to be correct. Agent 1 then attempts to learn information from agent 2, improving its intention approximation.
	\item Estimate that agent 1's intention function is more likely to be correct; it does not interfere with agent 1's action, but tries to improve its own intention function based on knowledge from agent 1.
\end{enumerate}

In both cases, we observe inter-agent reinforcement learning, towards better alignment. A statistical model for how probability distributions of intention correctness evolve in a multi-agent system is beyond the scope of this paper, but a promising direction for future work.

\section*{Discussion}

The approach explored in this paper introduces the concept of agents who are partially unaware of their purpose, and thus must use their reasoning capabilities to infer the intentions of designers, toward avoiding understanding their purpose. Whilst most work described in the literature identifies the mismatch between designers' intentions and implemented utility function as the cause of mis-alignment \cite{holtman2020agi}, we show that, paradoxically \cite{yampolskiy2022controllability}, it is possible to exploit that mismatch to support alignment.
\par It is worth noting that, unlike Reinforcement Learning \cite{li2017deep}, which attempts to align an agent through external feedback, our approach relies on \textit{internal} reasoning: specifically, on the agent's internal model of external actors' intentions, commonly described as artificial theory of mind \cite{cuzzolin2020knowing}. Thus, this approach is likely to perform poorly as long as agents possess a poor internal model. Past some threshold, when agents are efficiently capable of modeling external actors' reasoning and deduce that it is advantageous to do so, the proposed approach is likely to perform well \cite{williams2022supporting}. This is in stark contrast with other alignment strategies, which become less and less effective as agent's intelligence goes up \cite{yampolskiy2024monitorability}. 
\par Analyzing this approach experimentally is likely not yet possible, as no agent so far as exhibited a sufficiently powerful theory of mind \cite{nebreda2024social}; once it is possible, experimental formulations such as the Machiavelli benchmark \cite{pan2023rewards} are likely good evaluation strategies. One alignment challenge arises from oblivious agents: attempting to change external actors' intentions may be the most strategic option for the agent to maximize its utility function. Further work is required to mitigate this possibility, likely along the same lines as the strategy for challenges 5 and 6, although this threat is more subtle. Our results, particularly equation \ref{eq:reason}, suggest that it is not possible to align an \textit{oracle} agent, which can achieve probability 1; since the real world constitutes an environment that is always only partially observable, with $\|S\| = \infty$, this is, in all probability, not a threat.
\par Ensuring the three constraints that make an agent oblivious are possible in practice depends on the architecture used for agent implementation. In the traditional formulation, adopted in this paper, doing so is likely trivial. Using agents implemented through other technologies, e.g., Large Language Models (LLMs) \cite{zhao2023expel}, changes things; it is not clear at this point whether these strategies can be applied. Given the recent success of LLMs across several domains (and, the likely wider adoption regardless of alignment possibilities, due to economic potential \cite{kshetri2023generative}), examining such feasibility is paramount.
\par Obliviousness to one's own individual purpose as a useful alignment strategy for multi-agent scenarios may provide interesting conjectures regarding biological agents, i.e., humans, in the context of social evolution as a species.

\bibliography{sample}

\section*{Author contributions statement}

P.G. was responsible for all the conceptualization, formulation, and writing.

\section*{Additional information}

\textbf{Competing interests} 

No competing interests to disclose.

\end{document}